\documentclass[10pt,letterpaper]{article}

\usepackage{cogsci}

\cogscifinalcopy 

\usepackage{pslatex}
\usepackage{apacite}
\usepackage{float} 

\usepackage{amsmath,amssymb,amsfonts}
\usepackage{algorithmic}
\usepackage{graphicx}
\usepackage{textcomp}
\usepackage{xcolor}

\usepackage{eurosym,mathrsfs,multicol,indentfirst,color,bm,upgreek,booktabs}
\usepackage{tabularx}
\usepackage{subcaption}
\usepackage{CJK}
\usepackage{multirow}
\usepackage{makecell}
\usepackage{amssymb}
\usepackage{caption}

\usepackage{algorithm} 
\usepackage{algorithmic}
\usepackage{tablefootnote}
\usepackage{color,soul} 
\usepackage{tabu}
\usepackage{subfig,float}

\title{PiPMRE: A Pipeline Based on Language Model for Medical Relation Extraction}

\renewcommand{\thefootnote}{\fnsymbol{footnote}}
\author{
  {\large \bf Jiaxin Duan (duanjx@stu.pku.edu.cn)} \\
  School of Software and Microelectronics, Peking University, Beijing, China \\
  \AND {\large \bf Fengyu Lu (fengyul@stu.pku.edu.cn)} \\
  School of Software and Microelectronics, Peking University, Beijing, China \\
  \AND {\large \bf Junfei Liu (liujunfei@pku.edu.cn)} \\
  School of Software and Microelectronics, Peking University, Beijing, China \\ 
}

\begin{document}

\maketitle

{
  \renewcommand{\thefootnote}{}
  \footnotetext[99]{\textit{Preprint}: This work is accepted by the \textit{Proceedings of the 47th Annual Conference of the Cognitive Science Society}.}
}

\begin{abstract}
Medical relation extraction (MRE) is commonly known for extracting entities and their relations jointly from a medical text, which has attracted considerable attention in recent years.
Previous studies treat MRE as a sequence tagging task, which results in either a challenging design of the tagging schema or a failed extraction of multiple relations - due to intricate relationships among medical entities.
In this work, we review the task from the linguistic perspective and propose a novel pipeline framework, PiPMRE, developed on language models to enhance MRE performance. 
Specifically, PiPMRE consists of a relation generator and a relation filter. Given a text, the generator first yields multiple relational triplets, and then the filter scores each triplet and retains only those that pass the borderline as the final results. 
Implementing PiPMRE requires no tagging schema; instead, we use a simple template to reformulate the input text while ensuring entities and relations are generated in contextual order.
Extensive experimental results on two public datasets demonstrate the advancement of PiPMRE. 
It surpasses the previous state-of-the-art by an average of 5.6 recall points and 4.4 accuracy points. PiPMRE's superiorities are also demonstrated in few-shot settings.

\end{abstract}

\makeatletter

\section{Introduction}
\label{sec:1}
Medical relation extraction (MRE) aims to jointly extract paired entities and their relations from unstructured medical text into \textit{relation triplets} formatted like $<$\textit{subject entity, relation, object entity}$>$~\cite{YangHF23,Coling22}.
It is a foundation step in building knowledge-intensive applications, such as medical dialog systems~\cite{MedDia}, medical knowledge graph completion~\cite{MedKGC}, medical question answering~\cite{MedQA}, etc., and has attracted increasing attention in recent years.

\begin{figure*}[!tbp]
\centering
\includegraphics[width=0.9\linewidth]{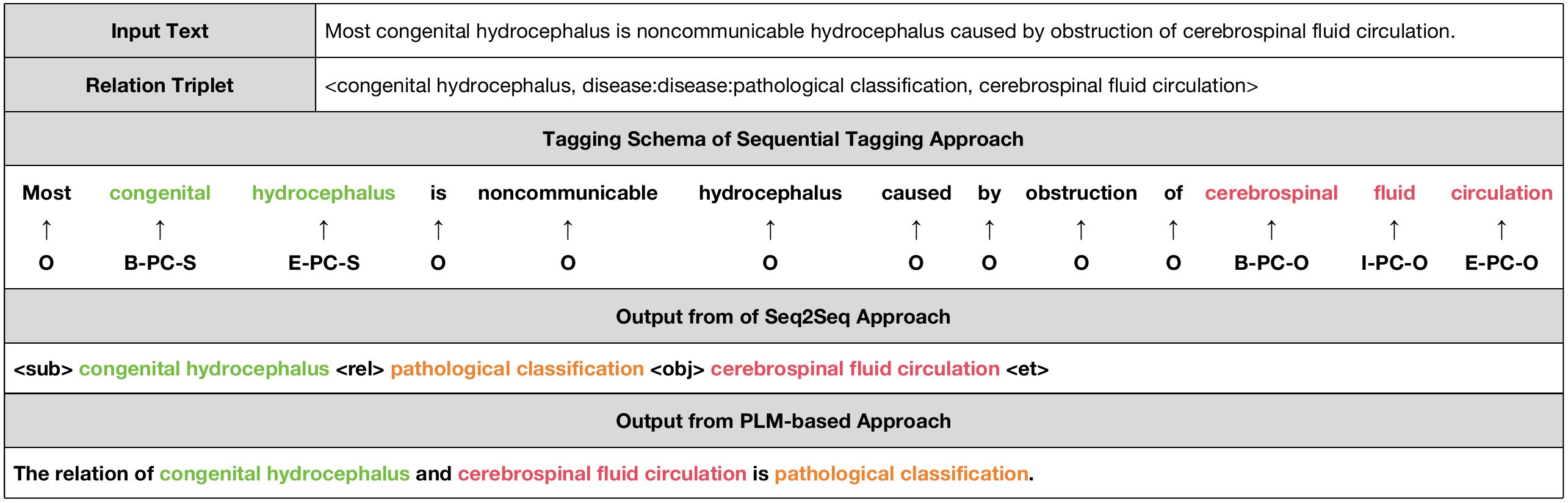}
\caption{
Comparison of MRE approaches in different styles.
Row four shows the tagging schema proposed in~\cite{ACL2017}, row six shows the linearized relation triple in~\cite{E-REBEL}, and the last row shows an example of triple textualization.
PC: pathological classification.
}
\label{fig:1}
\vspace{-10pt}
\end{figure*}

Existing approaches for MRE are mainly categorized into two groups, treating the task as a sequential or a sequence-to-sequence (Seq2Seq) tagging problem. 
As illustrated in Fig.~\ref{fig:1}, sequential tagging approaches~\cite{ACL2017,CASREL,TP-Linker,BiTT} put major effort into the design of tagging schema. In the fundamental work~\cite{ACL2017}, researchers imitate the BIEO notations used in named entity recognition (NER)~\cite{NER} and propose BIEO-R-SO tagging, where labels indicate the  \underline{\textbf{B}}egin/\underline{\textbf{I}}nner/\underline{\textbf{E}}nd token of a \underline{\textbf{S}}ubject/\underline{\textbf{O}}bjective entity in an inter-entities \underline{\textbf{R}}elation. Subsequently, \citeA{CASREL} leverage groups of binary sequences to tag overlapped relations, and \citeA{BiTT} introduce a binary tree to tag tree-form relational structures in the medical text. 
Since sequential tagging is a long-term topic in the machine learning community, this line of methods benefits from reusing mature technologies. However, mapping model-predicted tags to relational triplets is complicated, and how to tag an overwhelming number of terminologies and intricate relations in the medical domain remains an open question.
Seq2Seq approaches~\cite{GenIE,E-REBEL} instead learn to generate a linearized relation triplet conditioned on the given text, where subject, object, and relation are ordered by their contextual positions. They have advanced since giving up cumbersome tagging labor, but they are limited in multi-relation extraction due to the undetermined length of the target sequence~\cite{BiTT}.

In the open domain, recently proposed approaches focus on pre-trained language models (PLMs)~\cite{bert,t5} and convert relation extraction (RE) into a fundamental linguistic problem to facilitate evoking the parametric knowledge of PLM. \citeA{FPC} covert RE into masked language modeling, \citeA{GenPT} convert RE into text-denoising, and \citeA{UIE,TANL} convert RE into the generation of structured extraction language. 
However, despite their advantages in a short-sentence context, few previous studies have explored their potential in more complicated scenarios, especially in MRE.

In this paper, we propose PiPMRE, a novel PLM pipeline to improve current MRE performance. 
Specifically, PiPMRE consists of a relation generator and a relation filter. The generator is powered by a Transformer pre-trained with Seq2Seq text-infilling~\cite{t5}. To match its pre-training task, we rewrite an MRE instance, including the text and relational triplets, into a corrupted text and let the generator restore the corrupted words about entities and relations into another formatted text. 
When the pipeline works, the generator outputs a cluster of such texts via beam searching~\cite{beam-search}, and then the filter scores and compares each text with a borderline. The ones with a score above the borderline are accepted and mapped back to the final relation triplets.
Therefore, PiPMRE can not only extract an unknown number of relations into a fixed-length text but also utilize semantic information discarded in most previous studies. 

Beyond structuralism, we learn the generator in two continuous stages, which first injects medical knowledge into the model with incremental cross-domain pre-training and further tunes it for our reformulated MRE by preference optimization~\cite{DPO}. We also follow contrastive learning~\cite{slic} and train the filter with pairwise margin loss, ensuring that it understands the meaning of the borderline score. 
Through the generate-filter workflow and individual component learning, PiPMRE can adaptively decide multiple relation triplets for a given medical text without designing complex tagging schema or concerning about triplets overlapping.

Our main contributions are as follows:
\begin{itemize}
\item We propose a novel pipeline approach for MRE, which exhibits advantages over the previous methods on multi-relation extraction, as well as knowledge and semantic awareness.
\item We propose learning paradigms tailored to the generator and filter of PiPMRE, respectively.
\item Extensive experimental results on two public MRE datasets show that PiPMRE significantly outperforms the previous methods in both full-data and few-shot settings. We also conduct ablation studies to test the key factors that affect PiPMRE's performance.
\end{itemize}

\section{Methodology}

\subsection{Task Formulation}
\label{sec:3-1}
Given an MRE dataset $\mathcal{D}=\{\mathcal{X},\mathcal{T},\mathcal{E}\}$, where the $i$-th instance is a text (or sentence) $x_{i}\in \mathcal{X}$ containing $n_i$ medical entities $e_1,e_2,\cdots,e_{n_i}\in \mathcal{E}$, and the $j$-th pair of entities $e^{s}_{j},e^{o}_{j}$ has $m_j$ relations $r_1,r_2,\cdots,r_{m_j} \in \mathcal{T}$, the task of MRE is to extract all triplet $<e^{s}_{j},r_k,e^{o}_{j}>,i\in(0,|\mathcal{D}|),j\in (0,n_i),k\in (0,m_j)$ from $x_{i}$. 
Mathematically, this equals modeling the summed conditional probability:
\begin{equation}
\label{eq:1}
\sum_{i=1}^{|\mathcal{D}|}\sum_{j=1}^{n_i} \sum_{k=1}^{m_j}\mathrm{P}(<e^{s}_{j},r_k,e^{o}_{j}>|x_i).
\end{equation}

In solving this problem, we do not enumerate all possible triples $<e^{s}_{j},r_k,e^{o}_{j}>$ like work~\cite{ACL2017}, which is trivial and memory-consuming; also, we give up linearizing a triplet to a word sequence~\cite{GenIE} because its length varies on different instances. Instead, we textualize an instance into natural language and then corrupt the resulting instance to convert MRE as a text-infilling problem easily tackled with PLM. 

\textbf{Triplet Textualization.}
We use a simple language schema to express the relation triplet involved in MRE instances, where the entity type is also considered:
\begin{quote}
\centering
\scriptsize
\ The\ \{To\}\ \{Obj\}\ is\ the\ \{Rel\}\ of\ the\ \{Ts\}\ \{Sub\}.
\end{quote}
In this template, $\{Sub\},\{Obj\}$ are subject and object entities, $\{Ts\},\{To\}$ are the types of subject and object, respectively, and $\{Rel\}$ is their relation, e.g.,
\begin{quotation}
\scriptsize
$<$Congenital Hydrocephalus, pathological classification, Cerebrospinal fluid circulation$>$ $\rightarrow$ \textit{The disease cerebrospinal fluid circulation is the pathological classification of the disease congenital Hydrocephalus}.
\end{quotation}

\textbf{Instance Reformat.}
We then append the textualized triplet to the instance text. Consequently, an MRE instance is written in a coherent language context, easily understood by humans and language models, e.g.,
\begin{quotation}
\scriptsize
\textit{Most congenital hydrocephalus is non-communicable hydrocephalus caused by obstruction of cerebrospinal fluid circulation. The circulation of cerebrospinal fluid in the disease is the pathological classification of congenital hydrocephalus disease.}
\end{quotation}

\textbf{Instance Corruption.}
Finally, we make corruptions in the reformatted instance. Following~\cite{t5}, we replace the token spans located in the entities, entity types, and relation slots with distinct sentinel tokens while keeping the remaining content integral. A processed instance like the below schema:
\begin{quote}
\centering
\scriptsize
\{Text\}.\ The\ [TO]\ [OBJ]\ is\ the\ [R]\ of\ the\ [TS]\ [SUB]
\end{quote}
where [OBJ], [SUB], [TO], [TS], and [R] are sentinel tokens. 

On the above foundations, we approach MRE as Seq2Seq text-infilling, which conditions on a corrupted instance $\tilde{x}$ to predict a group of missed slots $y$ according to the sentinel tokens indication. $y$ is a structured text consisting of five slots with their contextual order in $\tilde{x}$: 
\begin{quote}
\centering
\scriptsize
[TO]\ \{To\}\ [OBJ]\ \{Obj\}\ [R]\ \{Rel\}\ [TS]\ \{Ts\}\ [SUB]\ \{Sub\}
\end{quote}

\begin{figure*}[!htbp]
\centering
\includegraphics[width=0.95\textwidth]{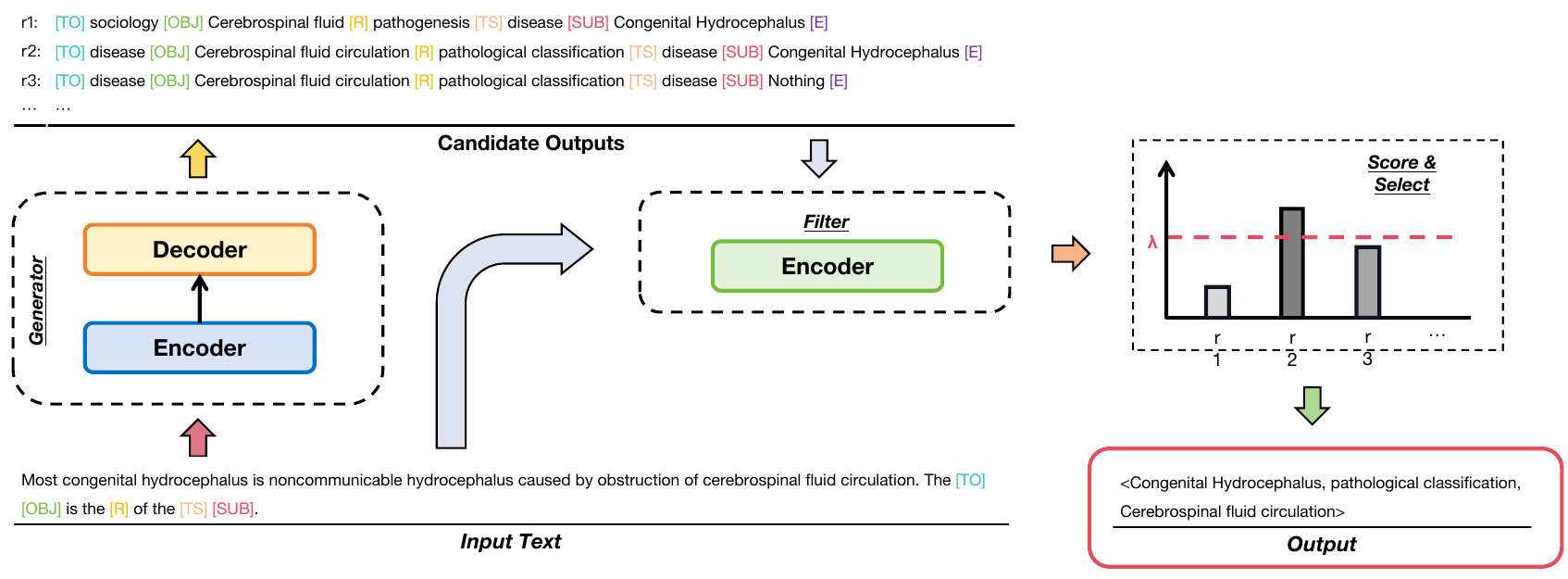}
\caption{
The overview of our PiPMRE.
$\{\cdot\}$ locates the slot to be filled, [$\cdots$] is a special token, and [E] is added to the text end.
}
\label{fig:2}
\end{figure*}

\subsection{Relation Generator}
\label{sec:3-2}
We start the PiPMRE generator from T5~\cite{t5} - a Transformer model pre-trained with Seq2Seq text-infilling and has shown significant strength on extractive linguistic tasks.
Additionally, we perform an incremental cross-domain pre-training (ICPT) to inject medical knowledge into the model and then fine-tune it toward our task requirements.

\textbf{Incremental pre-training.} Given an entity $e$ in the medical knowledge graph $\mathcal{G}$, its \textit{neighbors} consists of entities linked $e$ by a direct relation and \textit{indirect neighbors} $\bar{\mathcal{E}}_{i}$ link $e$ by 2-skip relations. 
A corpus entailing medical knowledge is easily created on this assumption.
For each $e_{i}\in \mathcal{G}$, we ask Llama3.1~\cite{llama3} to make a sentence conditioned on one of its neighbors $\bar{e}_{i}$ and one of its indirect neighbors $\tilde{e}_{i}$ by instructing it with the following prompt:
\begin{quotation}
Considering the $\bar{e}_{i}$ is $r_{1}$ of $e_{i}$ and $\tilde{e}_{i}$ is $r_{2}$ of $e_{i}$, please make a sentence using $e_{i}$, $\bar{e}_{i}$ and $\tilde{e}_{i}$.
\end{quotation}
We mask entity tokens in the text by probability 0.8 while other tokens by 0.2. Then, we train the generator to restore the original text at the target end.

In this work, we build our pre-training corpus with CMeKG\footnote{http://cmekg.pcl.ac.cn/}, which widely covers 6,310 diseases, 19,853 drugs, 1,237 diagnosis and treatment technologies, and more than 30 common relationships. The final corpus contains 269,930 sentences, and the supervised cross-entropy loss is used during the training process.



\textbf{Fine-tuning.} During fine-tuning, the biggest challenge is to preserve pre-training knowledge. 
Thanks to prompt tuning~\cite{prompt-tuning} technology, which fine-tunes a PLM while freezing all pre-trained parameters, this problem is alleviated to a great extent. 
Noting the generator intakes a corrupted instance $\tilde{x}$ and aims to output the target text $y$ containing the slots missed in $\tilde{x}$ and joined by sentinel tokens. It models the following probabilistic:
\begin{equation}
\label{eq:2}
\begin{aligned}
\mathrm{P}(y|\tilde{x};\theta)&=\prod_{t=1}^{|y|} \mathrm{P}(y_{t}|\tilde{x},y_{<t};\theta)\\
\mathrm{P}(y_{t}|\tilde{x},y_{<t};\theta)&=Dec(Enc(\tilde{x};\theta ),y_{<t};\theta)
\end{aligned},
\end{equation}
where $\theta$ denotes the generator's parameters, and $Enc,Dec$ are the generator's encoder and decoder, respectively.
Following~\cite{prompt-tuning}, we insert $n$ continuous soft tokens $\mathbf{c}=c_1,c_2,\cdots,c_n$ at the beginning of $\tilde{x}$, and then Eq.~\ref{eq:2} converts to:
\begin{equation}
\label{eq:3}
\begin{array}{c}
\mathrm{P}(y|\tilde{x};\theta)\to \mathrm{P}(y|c_{1:n},\tilde{x};\theta,\phi ) \\
\end{array},
\end{equation}
where
\begin{equation}
\label{eq:4}
\mathrm{P}(y_t|c_{1:n},\tilde{x},y_{<t};\theta,\phi)=Dec(Enc([\mathbf{c};\tilde{x}];\theta),y_{<t};\theta,\phi),
\end{equation}
$\phi$ denotes the embeddings of $\mathbf{c}$, which are learnable, and $[\cdot;\cdot]$ means texts connection.


Besides, to meet the intricate medical scenario, where more than one relation triplet may be extracted from a text, we introduce an additional objective, which aims to maximize the probability of texts $y^{+}\in \mathcal{T}^{+}_{x}$ established by gold relation triplets while minimizing the ones $y^{-}\in \mathcal{T}^{-}_{x}$ built on fake triplets:
\begin{equation}
\label{eq:7}
\max_{\phi } \mathbb{E}_{y^{+},y^{-}} [\mathrm{P}(y^{+}|\mathbf{c}, \tilde{x};\theta,\phi)-\mathrm{P}(y^{-}|\mathbf{c},\tilde{x};\theta,\phi)].
\end{equation}

After incremental pre-training, we follow direct preference optimization (DPO)~\cite{DPO} and continuously train the model to minimize the loss $\mathcal{L}_{dpo}(\phi)$:
\begin{equation}
\label{eq:8}
-\underset{y^{+},y^{-}}{\mathbb{E} } [\log \sigma\left(\beta \log \frac{\pi_{\phi^{*}}\left(y^{+} \mid \tilde{x}\right)}{\pi_{\theta^{*}}\left(y^{+} \mid \tilde{x}\right)}-\beta \log \frac{\pi_{\phi^{*}}\left(y^{-} \mid \tilde{x}\right)}{\pi_{\theta^{*}}\left(y^{-} \mid \tilde{x}\right)}\right)],
\end{equation}
where $\pi_{\phi^{}}(\cdot|\tilde{x})$ is an abbr. of $\mathrm{P}(\cdot|\mathbf{c}, \tilde{x};\theta,\phi)$, $\theta^{*}$ denotes a copy of model parameters learned with supervised fine-tuning, and $\phi^{*}$ means $\phi$ updated after back-propagation. 
The next section discusses the construction of fault target texts $\mathcal{T}_{x}^{-}$.

\subsection{Relation Filter}
\label{sec:3-3}
The PiPMRE filter is built upon BERT~\cite{bert} and estimates whether the generator infers correct triplet elements from a corrupted instance. Formally, given an instance text $x$, and the text $y^{*}\sim g_{\phi}(\tilde{x})$ sampled from the generator outputs $g_{\phi}(\cdot)$, the function of the filter is $f(x,y^{*})\to s, s \in (0,1)$.
Like what has performed on the generator, we embed $x$ and the slots in $y^{*}$ into a language template to match the filter's pre-training:
\begin{quote}
\centering
\scriptsize
\{Text\}.\ The\ \{Obj\}\ [M]\ (is\ or\ isn't) \ the\ \{Rel\}\ of\ the\ \{Sub\}.    
\end{quote}

It is seen that we abandon the entity-type slots in $y^{*}$, which are proven negligible in our preliminary experiments, and require the filter to predict the mask token [M] - \textit{is} or \textit{isn't}. 
Practically, the filter models the probability $\mathrm{P}(\cdot|T[x,y^{*}])$, where $T[\cdot,\cdot]$ means template infilling, rather than making an absolute binary choice between the two given words. To bridge this gap, we sharpen the filter's output distribution:
\begin{equation}
\label{eq:9}
\mathrm{\hat{P}}(v_{i} |T[x,y^{*}])=\frac{\exp \left(\bar{v}_{i} /\tau\right)}{\sum_{k}^{N} \exp \left(\bar{v}_{k} /\tau\right)},
\end{equation}
where $\tau<1$ is a temperature and $\bar{v}_{i}$ is the output of the filter's second last layer corresponding to the word $v_{i}$. Furthermore, we let $f(x,y^{*})=\mathrm{\hat{P}}(is |T[x,y^{*}])$ quantify the confidence of using the entity and relation slots $\{Sub\}^{*},\{Rel\}^{*},\{Obj\}^{*}$ in $y^{*}$ to compose a correct relation triplet that belongs to $x$.

Based on the above mentions, we learn the filter to provide reasonable ratings, depending on the validness of the generator output. Feeding in an instance $x$, the generator's output text is either made up of elements in gold relation triplet or not. We collect the former into $\mathcal{T}_{x}^{+}$ and the later $\mathcal{T}_{x}^{-}$. Intuitively, a negative case $y^{-}_{i}\in \mathcal{T}_{x}^{-}$ is easily fabricated by replacing one or more slots in $y^{+}_{i}\in \mathcal{T}_{x}^{+}$ with random tokens or peer tokens stick with entity/relation/entity type.
We train the filter with a cloze prompt tuning approach, i.e., inserting soft tokens $t_1,t_2,\cdots,t_m$ (with learnable embeddings $\psi$) after the \{\textit{Text}\} slot in the filter input text, also, a pair-wise margin loss $\mathcal{L}_{ctl}(\psi)$ is used inspired by contrastive learning~\cite{slic}:
\begin{equation}
\label{eq:10}
\sum_{i=1}^{|\mathcal{T}_{x}^{+}|} \max (\zeta-f_{\psi }(x,y^{+}_{i}),0)+\max (f_{\psi }(x,y^{-}_{i})-\zeta,0).
\end{equation}
Notably, $\mathcal{T}_{x}^{-}$ is dynamically derived from $\mathcal{T}_{x}^{+}$ during training, depending on the instance $x$, therefore, the two sets have an identical size. $\zeta$ is a borderline parameter. 

\subsection{Pipeline Inference}
\label{sec:3-4}
The inference of PiPMRE is described in Fig.~\ref{fig:2}. Starting from a templated instance, the generator first produces a cluster of candidate outputs, then, the filter scores and compares each with a borderline $\zeta$ to decide its correctness.

\textbf{Modulated Decoding.}
Due to the randomness of autoregressive generation, each token in the generated target text is conditionally sampled from the generator vocabulary, causing a risk of yielding invalid slots that fail to match any relation triplet. 
To address this problem, we draw ideas of \cite{GenIE,GenPT} and propose a \textit{modulated decoding} mechanism. 

Given a corrupted instance $\tilde{x}^{*}$ to the encoder and a starting token [TO] to the decoder, the generator adopts beam search (BS)~\cite{beam-search} to sample $K$ candidate target texts $y^{*}_{1},y^{*}_{2},\cdots,y^{*}_{K}$.
In this process, the searching space of BS is constrained by a Trie~\cite{SURE}, where the possible next token in a branch can only be one of the children of the last visited node, depending on the pre-given sets $\mathcal{T}$ and $\mathcal{E}$. Taking the instance in Fig~\ref{fig:2}, once the model has generated an incomplete phrase \textit{[TO] disease [OBJ]}, the next token must start an entity of the type disease; also, the next token of the phrase \textit{[TO] disease [OBJ] Cerebrospinal fluid [R] pathogenesis [TS]} must be \textit{disease} because only a disease can be the object of \textit{pathogenesis}.
This strategy offers higher efficiency than searching on the entire vocabulary and ensures the validness of model outputs \cite{GenIE}. 

\textbf{Scoring and Filtering.}
Finally, the filter scores each of the generator's outputs: 
\begin{equation}
\label{eq:11}
s_{i}=f_{\psi }(x,y^{*}_{i}),i=1,\cdots, K.
\end{equation}
Each $s_{i}$ is compared with the borderline $\zeta$. If $s_{i}>\zeta$, the triplet $<\{Sub\}_i^{*},\{Rel\}_i^{*},\{Obj\}_i^{*}>$, is deemed established for the instance $x^{*}$ ($\{\cdot\}_i^{*}$ is the slot in $y_{i}^{*}$); otherwise, it is abandoned.

\begin{table}[!ht]
\centering
\scriptsize
\caption{
Datasets Statistics.
}
\begin{tabular}{ccccccc}
\toprule
\multirow{2.5}{*}{\textbf{Dataset}} & \multicolumn{2}{c}{\textbf{\#Instance}} & \multicolumn{2}{c}{\textbf{\#Relations}} & \multicolumn{2}{c}{\textbf{\#RPI}} \\
\cmidrule(lr){2-3}\cmidrule(lr){4-5}\cmidrule(lr){6-7}
 & \textbf{Train} & \textbf{Test} & \textbf{Train} & \textbf{Test} & \textbf{Train} & \textbf{Test} \\
\midrule
CHIP & 14,339 & 3,585 & 43,660 & 10,626 & 3.04 & 2.95 \\
CMeIE & 14,339 & 3,585 & 64,835 & 16,309 & 1.14 & 4.55 \\
\bottomrule
\end{tabular}
\label{tab:1}
\vspace{-10pt}
\end{table}

\section{Experiments}
\subsection{Datasets}
\label{sec:5-1}
\textbf{CMeIE}~\cite{2020CMeIE} and \textbf{CHIP}~\cite{CHIP} are used to evaluate our method.
Table~\ref{tab:1} reports their statistics.


\subsection{Comparison Methods}
We compare our PiPMRE with nine advanced RE methods scattered in three types.
\underline{\textit{Sequential tagging methods:}} 
\textbf{NovelTagging}~\cite{ACL2017} proposes a BIEO-R-SO tagging schema for relation extraction.
\textbf{CASREL}~\cite{CASREL} extends the tagging label from a single numerical sequence to groups of binary sequences to cope with overlapped relations.
\textbf{TP-Linker}~\cite{TP-Linker} extracts entities and relations simultaneously using a novel handshaking tagging strategy.
\textbf{BiTT}~\cite{BiTT} develop binary tree-based tagging schema to capture tree-like relation structure in medical texts.
\underline{\textit{Seq2Seq methods:}} 
\textbf{GenIE}~\cite{GenIE} generates linearized relation triplet from a given sentence, where special tokens $<$sub$>$, $<$rel$>$, and $<$obj$>$ demarcate the start of a subject, relation, and object, respectively, and $<$et$>$ demarcates the object end.
\textbf{E-REBEL}~\cite{E-REBEL} uses $<$triplet$>$ to mark the start of a relation triplet and $<$sub$>$ and $<$obj$>$ to split subject, object, and their relation. 
\underline{\textit{PLM-based methods:}} 
\textbf{FPC}~\cite{FPC} transforms RE into MLM and fine-tunes a BERT model with curriculum-guided prompting.
\textbf{TANL}~\cite{TANL} frames RE as a translation task based on augmented natural language.
\textbf{GenPT}~\cite{GenPT} converts RE to a Seq2Seq text-denoising task, recovering the corrupted entities and their relation at the target end.

\begin{table}[!t]
\centering
\scriptsize
\caption{
Evaluation results on CHIP test set. 
\underline{Underline} results are the previous best.
\textbf{Bold} results are the best.
}
\begin{tabular}{c|c|cp{20pt}<{\centering}p{20pt}<{\centering}}
\toprule
\textbf{Method} & \textbf{Base Model} & \textbf{Precision} & \textbf{Recall} & \textbf{F1}\\
\midrule
\midrule
\multicolumn{5}{c}{\textbf{Sequential Tagging Methods}} \\
\midrule
\midrule
NovelTagging & BERT-\textit{base} & 74.6 & 71.0 & 71.1 \\
CASREL & BERT-\textit{base} & 80.1 & 71.8 & 71.8 \\
TP-Linker & BERT-\textit{base} & 83.1 & 68.5 & 77.4 \\
BiTT & BERT-\textit{base} & 82.1 & 73.2 & 84.4 \\
\midrule
\midrule
\multicolumn{5}{c}{\textbf{Seq2Seq Methods}} \\
\midrule
\midrule
GenIE & BART-\textit{large} & 82.1 & 73.2 & 80.4 \\ 
E-REBEL & REBEL-\textit{large} & 81.4 & 68.3 & 80.9 \\ 
\midrule
\midrule
\multicolumn{5}{c}{\textbf{PLM-based Methods}} \\
\midrule
\midrule
FPC & RoBARTa-\textit{large} & 83.1 & 68.5 & 77.4 \\
TANL & T5-\textit{large} & 82.4 & 74.0 & 78.0 \\
GenPT & BART-\textit{large} & \underline{89.2} & \underline{84.0} & \underline{87.8} \\
\midrule
PiPMRE & T5-\textit{large} & \textbf{89.7} & \textbf{86.1} & \textbf{88.9} \\
\bottomrule
\end{tabular}
\label{tab:2}
\vspace{-10pt}
\end{table}

\subsection{Implementation Details}
In our experiments, we implement the PiPMRE generator with T5-\textit{large}~\footnote{https://huggingface.co/IDEA-CCNL/Randeng-T5-784M} and implement the filter with BERT-\textit{base}~\footnote{https://huggingface.co/google-bert/bert-base-chinese}. 
As for hyperparameters, we set $\gamma=0.2$, $\beta=0.6$, $\tau=0.2$ and $\zeta=0.45$. Besides, we set the length of soft tokens used in prompt tuning to 20 for both the generator and filter. When beam search~\cite{beam-search} is used, the beam width $K$ is set to 16. 
Our codes are built with Pytorch\footnote{https://pytorch.org} and Huggingface Transformer\footnote{https://huggingface.co/models} libraries, and 8/4 NVIDIA RTX 4090 GPUs are used to run model training/inference. 
We incrementally pre-train the PiPMRE generator for 10K steps on our collected corpus, the batch size is 32, and an AdamW~\cite{adamW} optimizer with a learning rate 1e-4 is used. 
During fine-tuning, we train the PiPMRE's two components for at most 10 epochs. We bind the AdamW optimizer with a linear learning schedule, where the initial learning rate is 2.5e-5 for the generator and 1e-4 for the filter. It warms up during the first 10\% training steps and decays to 0 gradually in the subsequent steps.


Following previous works, we take micro precision (\%), recall (\%), and F1 score (\%) as the main metrics for model evaluation. 
Besides, our experiments also compare the foundation model (and its scale) used in each method (the tagger network in tagging methods, the Seq2Seq model used in Seq2Seq methods, and the language model used in PLM-based methods).

\begin{table}[!t]
\centering
\scriptsize
\caption{
Evaluation results on CMeIE test set. 
Results with ${\dag}$ are reported in the original or previous papers, otherwise from our reproduction. 
}
\begin{tabular}{c|p{45pt}<{\centering}|cp{20pt}<{\centering}p{20pt}<{\centering}}
\toprule
\textbf{Method} & \textbf{Parameters} & \textbf{Precision} & \textbf{Recall} & \textbf{F1}\\
\midrule
\midrule
\multicolumn{5}{c}{\textbf{Sequential Tagging Methods}} \\
\midrule
\midrule
NovelTagging & 110M & 51.4$^{\dag}$ & 17.1$^{\dag}$ & 25.6$^{\dag}$ \\
CASREL & 110M & 53.5$^{\dag}$ & 28.2$^{\dag}$ & 37.0$^{\dag}$ \\
TP-Linker & 110M & 52.3 & 27.7 & 38.5 \\
BiTT & 110M & \underline{55.6}$^{\dag}$ & \underline{45.5}$^{\dag}$ & \underline{50.1}$^{\dag}$ \\
\midrule
\midrule
\multicolumn{5}{c}{\textbf{Seq2Seq Methods}} \\
\midrule
\midrule
GenIE & 406M & 51.7 & 27.7 & 38.5 \\ 
E-REBEL & 770M & 51.3 & 19.9 & 35.9 \\ 
\midrule
\midrule
\multicolumn{5}{c}{\textbf{PLM-based Methods}} \\
\midrule
\midrule
FPC & 355M & 52.6 & 39.3 & 44.5 \\
TANL & 770M & 53.3 & 33.9 & 46.0 \\
GenPT & 406M & \underline{55.6} & 43.1 & 49.6 \\
\midrule
PiPMRE & 700M & \textbf{56.7} & \textbf{46.6} & \textbf{50.3} \\
\bottomrule
\end{tabular}
\label{tab:3}
\vspace{-10pt}
\end{table}

\subsection{Main Results}
Table~\ref{tab:2} and Table~\ref{tab:3} present the comprehensive evaluation results of our PiPMRE and the comparison methods. 
On both datasets, BiTT and GenIE exhibited individually the best performance in their group. GenPT outperformed them, and our PiPMRE further surpassed GenPT by 1.1 points of F1 on CHIP and 0.7 points of F1 on CMeIE. We note that GenPT converts MRE as a Seq2Seq text-denoising problem, similar to ours. To some extent, this suggests the compatibility between the two tasks, indicating a promising direction for future studies. 
On the other hand, the methods mentioned in our experiments can also be grouped according to their base model. Taking the BERT group as an example, the success of BiTT indicated the importance of exploiting the semantic structure of medical text. With BART-\textit{large} as the backbone, GenPT showed significant superiorities over GenIE. GenPT performed a task transformation that exactly meets the BART pre-training purpose, i.e., text-denoising. On the contrary, GenIE only uses BART's function of Seq2Seq generation, without matching pre-training and downstream tasks or utilizing the model's linguistic ability. Therefore, PLM-based approaches present an adequate idea that directs pre-trained Transformers for RE-like tasks.
Finally, our PiPMRE performed better than TANL. Consider that T5 is pre-trained with text-infilling while TANL converts RE to a machine translation task. This comparison highlights again the consistency of pre-training and downstream tasks.

\begin{table}[!t]
\centering
\scriptsize
\caption{
Evaluation results on CHIP and CMeIE test instances with single relation.
}
\begin{tabular}{ccccc}
\toprule
\multirow{2.5}{*}{\textbf{Method}} & \multicolumn{2}{c}{\textbf{CHIP}} & \multicolumn{2}{c}{\textbf{CMeIE}} \\
\cmidrule(lr){2-3}\cmidrule(lr){4-5}
 & \textbf{H@1} & \textbf{H@3} & \textbf{H@1} & \textbf{H@3} \\
\midrule
CASREL & 70.16 & 98.84 & 76.24 & 98.47 \\
BiTT & 80.41 & 99.05 & 81.28 & \textbf{99.82} \\
GenIE & 74.84 & 97.94 & 79.30 & 98.03 \\
GenPT & 86.86 & 99.01 & 85.64 & 99.34 \\
\midrule
PiPMRE & \textbf{91.08} & \textbf{99.65} & \textbf{92.14} & 99.39 \\
\textit{- w/o} DPO & 89.18 & 99.01 & 89.48 & 99.31 \\
\bottomrule
\end{tabular}
\label{tab:4}
\end{table}

\begin{table}[!t]
\centering
\scriptsize
\caption{
Evaluation results on CHIP and CMeIE test instances with multiple relations.
F1 scores are reported.
}
\begin{tabular}{ccccccc}
\toprule
\multirow{2.5}{*}{\textbf{Method}} & \multicolumn{3}{c}{\textbf{CHIP}} & \multicolumn{3}{c}{\textbf{CMeIE}} \\
\cmidrule(lr){2-4}\cmidrule(lr){5-7}
 & \textit{NEO} & \textit{EPO} & \textit{SEO} & \textit{NEO} & \textit{EPO} & \textit{SEO} \\
\midrule
CASREL & 81.4 & 81.1 & 80.1 & 47.1 & 45.8 & 46.1 \\
BiTT & 82.7 & 83.7 & 83.6 & 47.7 & \textbf{49.4} & \textbf{50.3} \\
GenIE & 79.8 & 67.1 & 74.8 & 36.3 & 42.3 & 43.1 \\
GenPT & 78.6 & 67.7 & 75.4 & 37.5 & 41.8 & 42.4 \\
\midrule
PiPMRE & \textbf{85.7} & \textbf{88.0} & \textbf{87.3} & \textbf{50.0} & \textbf{49.4} & 50.2 \\
\bottomrule
\end{tabular}
\label{tab:5}
\vspace{-10pt}
\end{table}

\subsubsection{Single-relation MRE}
To make the evaluation of single-relation MRE fair for various methods, we introduce a novel metric - H@M. 
Given an instance associated with a relation triplet $r$, we can sample top-$M$ candidate triplets $\mathcal{C}_M$ from a RE model's outputs according to estimated likelihood, and:
\begin{equation}
H@M=\mathbb{I}(r \subset \mathcal{R}_M ),
\end{equation}
where $\mathbb{I}(\cdot)$ is 1 if the condition in $(\cdot)$ holds, otherwise 0.
Methods in Table~\ref{tab:4} performed closely evaluated by H@3. 
However, when the metric converts to H@1, our PiPMRE showed obvious advantages. We attribute this to the fine-tuning of the PiPMRE generator, which enables the model to produce a more promising triplet by a higher probability. The last row of Table~\ref{tab:4} proved our view.

\subsubsection{Multi-relation MRE}
\citeA{BiTT,TP-Linker} recognize multi-relation RE into three scenarios: with a pair of entities overlapped (EPO), single entity overlapped (SEO), and no entity overlapped (NEO). Table~\ref{tab:5} reports the evaluation results of multi-relation RE\footnote{The CHIP/CMeIE test set contains 40/19 EOP instances and 2,054/3,000 SEO instances, respectively.}.
GenIE and GenPT showed weakness across all scenarios. Although we set the maximum length of the output sequence to 256 tokens, they tend to generate incomplete sequences or invalid ones that fail to recover relation triplets due to the \textit{repetition problem}. In contrast, sequential tagging approaches showed more promising results. However, in most cases, our PiPMRE surpassed the previous state-of-the-art BiTT by a large margin, one reason for its advanced comprehensive performance.

\begin{table}[!t]
\centering
\scriptsize
\caption{
Evaluation results of PiPMRE variants on CMeIE test set. The declined scores are reported.
CTL: contrastive learning.
$\triangle $: we keep the original setting of this component.
}
\label{tab2}
\begin{tabular}{c|c|ccc}
\toprule
\textbf{Generator} & \textbf{Filter} & \textbf{Precision} & \textbf{Recall} & \textbf{F1} \\
\midrule
\textit{- w/o} ICPT & $\triangle $ & $\uparrow$ 1.8 & $\uparrow$ 1.5 & $\uparrow$ 0.9 \\
\textit{- w/o} DPO & $\triangle $ & $\downarrow $ 2.9 & $\downarrow $ 1.6 & $\downarrow $ 2.1 \\
\midrule
$\triangle $ & \textit{- w/o} CTL & $\downarrow $ 3.7 & $\downarrow $ 4.5 & $\downarrow $ 3.8 \\
\bottomrule
\end{tabular}
\label{tab:6}
\vspace{-10pt}
\end{table}

\subsection{Further Analysis}
\subsubsection{Ablation Study}
To further analyze the importance of each technology used in building PiPMRE, we consider five variants of PiPMRE and compare their performance in Table~\ref{tab:5}.
When the filter is kept, the lack of DPO during the generator fine-tuning process impacts the PiPMRE performance most, followed by the absence of incremental pre-training.
Once we abandon contrastive learning for the filter, we assign label 1/0 to a text $y_{i}^{+}$/$y_{i}^{-}$ and supervised train the model with a cross-entropy loss. It greatly harms the PiPMRE performance because of inadequate scoring conducted on the generator outputs.



\subsubsection{Few-shot Learning} 
\begin{figure}[!t]
\centering
\begin{subfigure}{0.5\linewidth} 
\includegraphics[width=1.0\linewidth]{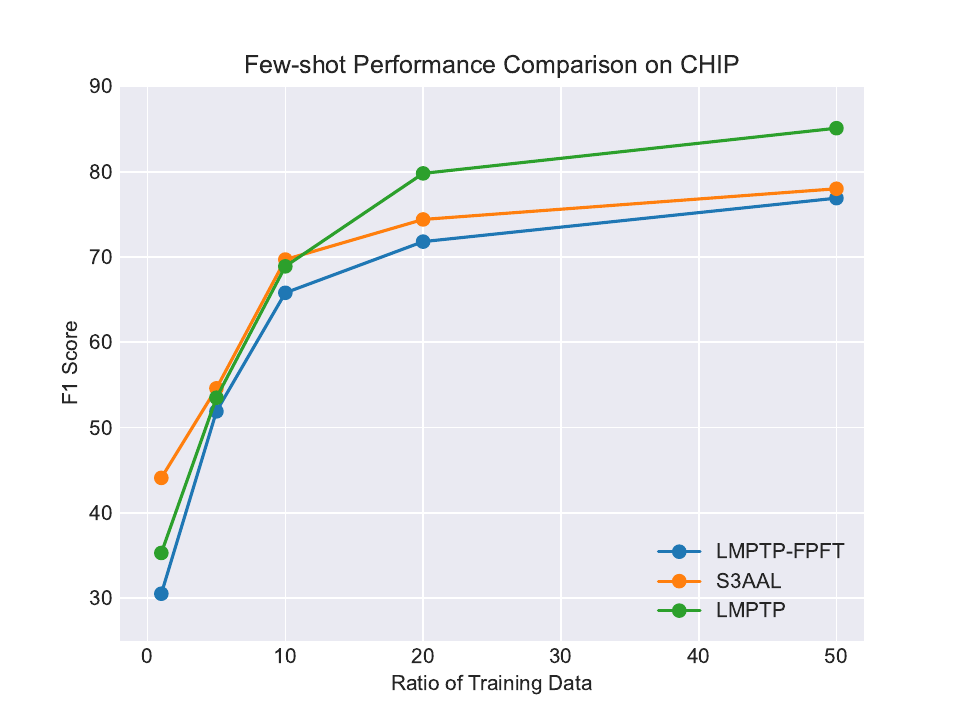} 
\caption{Few-shot MRE on CHIP.}
\label{fig:3a}
\end{subfigure}%
\begin{subfigure}{0.5\linewidth}
\includegraphics[width=1.0\linewidth]{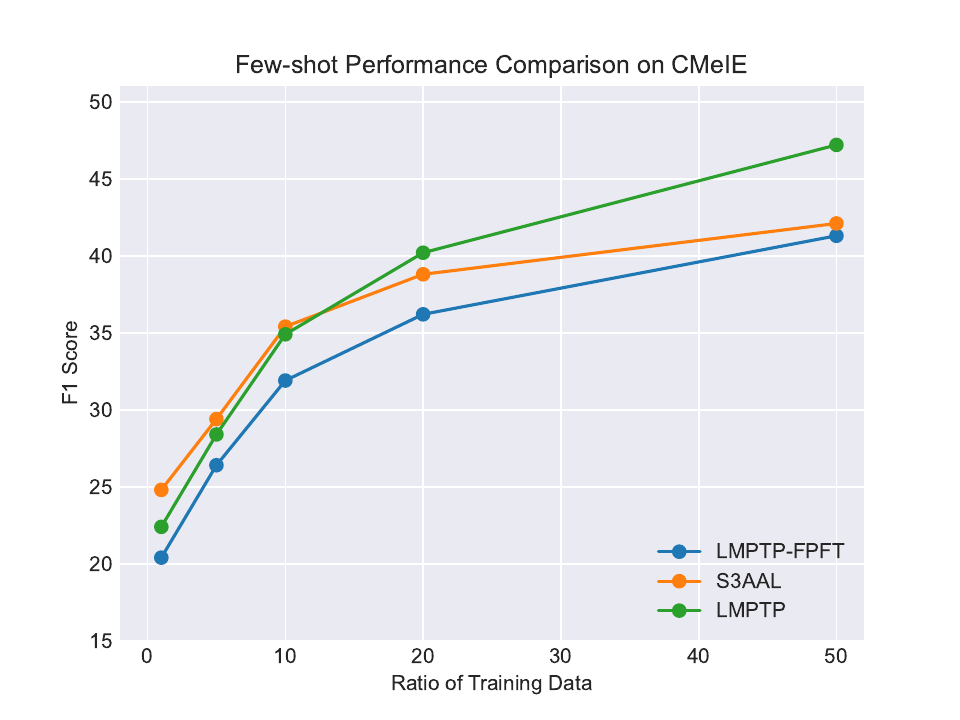} 
\caption{Few-shot MRE on CMeIE.}
\label{fig:3b}
\end{subfigure}
\caption{Few-shot performance on CHIP (a) and CMeIE (b).}
\label{fig:3}
\vspace{-10pt}
\end{figure}

We also test our method in few-shot settings to follow previous studies. 
We train PiPMRE (its generator and filter) and S$^{3}$AAL~\cite{S3AAL} on 1\%$\sim $50\% training samples (S$^{3}$AAL learns a support set, especially for few-shot MRE). 
Notably, since prompt tuning (PT) is famous for advancement in few-shot tasks, the PiPMRE variant that gives up PT in both components' learning process (referred to as PiPMRE-FPFT) is also considered. The comparison results are displayed in Fig.~\ref{fig:3}, evaluated by the micro F1 score. 
The original PiPMRE outperformed the varients without prompt tuning, no matter how many training samples were used. 
On the other hand, PiPMRE performed close to S$^{3}$AAL with less than 10\% training samples, and the gap between the two methods increases with the increased number of training samples.
This is because S$^{3}$AAL adapts only for few-shot scenarios, and the support set it learned shows limited advantages given enough training data.


\section{Conclusion}
\label{sec:7}
This paper introduces a novel two-stage pipeline for MRE, named PiPMRE. 
PiPMRE adopts a Seq2Seq PLM to generate formatted texts containing grouped elements of relational triplet and a filter to validate the generation results.
In contrast to traditional approaches that carry on sequence-based tagging, PiPMRE abandons the design of complicated tagging schema while adequately tackling medical texts containing multiple relation triplets.
Experimental results in diverse full-data/few-shot and single-/multiple-relation settings demonstrate the robustness of PiPMRE. 




\newpage
\bibliographystyle{apacite}

\setlength{\bibleftmargin}{.125in}
\setlength{\bibindent}{-\bibleftmargin}

\bibliography{references,custom}

\end{document}